Title: Efficient Force and Stiffness Prediction in Robotic Produce Handling with a Piezoresistive Pressure Sensor

Authors: Preston Fairchild*, Claudia Chen, Xiaobo Tan

*Corresponding author

Department of Electrical and Computer Engineering, Michigan State University, East Lansing, MI 48824, USA. Email: fairch42@msu.edu, chenclau@msu.edu, xbtan@msu.edu

Abstract: Properly handling delicate produce with robotic manipulators is a major part of the future role of automation in agricultural harvesting and processing. Grasping with the correct amount of force is crucial in not only ensuring proper grip on the object, but also to avoid damaging or bruising the product. In this work, a flexible pressure sensor that is both low cost and easy to fabricate is integrated with robotic grippers for working with produce of varying shapes, sizes, and stiffnesses. The sensor is successfully integrated with both a rigid robotic gripper, as well as a pneumatically actuated soft finger. Furthermore, an algorithm is proposed for accelerated estimation of the steady-state value of the sensor output based on the transient response data, to enable real-time applications. The sensor is shown to be effective in incorporating feedback to correctly grasp objects of unknown sizes and stiffnesses. At the same time, the sensor provides estimates for these values which can be utilized for identification of qualities such as ripeness levels and bruising. It is also shown to be able to provide force feedback for objects of variable stiffnesses. This enables future use not only for produce identification, but also for tasks such as quality control and selective distribution based on ripeness levels.

Keywords: Robotics, sensing, produce handling, grasping

Highlights:

- Low-cost and easy-to-fabricate sensor for easy implementation with a variety of robotic grippers
- Fast estimation of settled resistance using exponential decay curve fit
- Measurements of grasping force and stiffness of a held object
- Various produce handling features such as ripeness monitoring, bruising detection, and size estimation

1. Introduction:

The use of robotic end-effectors for securely grasping objects is a pivotal component in manipulation tasks. Some specifically designed tools used for harvesting purposes include a pineapple harvesting gripper with a cutting device [1], a robotic harvester to dig and cut radicchio [2], a cable-driven strawberry harvester which encompasses the fruit as it is harvested [3], a tomato harvester integrating force feedback [4], a tendon driven finger-like robotic apple harvester [5], and a robotic platform for grape harvesting [6]. Robotic planters, such as a rice planting robot [7], can have similarly specific designs. Robotic grippers designed for targeted applications generally increase the efficiency of their task but lose out on the capability to operate with other objects effectively or even entirely. For many applications, such as local distribution, multiple different types of produce may need to be handled, necessitating a more generic robotic gripper. Robotic graspers designed to manipulate a wide range of produce include configurable soft robotic fingers [8], a particle jamming grasper [9], a grasper with adaptively shaped fingers [10], a suction cup gripper with adaptive sealing [11], and a hybrid finger and particle jamming gripper [12]. While these grippers are more adaptable to a variety of produce, they also require additional care in terms of force and positioning control for specific items to not damage them. This necessitates the use of some kind of sensor to monitor the output of the grasper to avoid damage and handle a range of produce.

Successfully implementing a more generic gripper for a variety of use cases requires different control strategies. The force required to grasp an apple is significantly more than that of a strawberry, and if the same force is used for both, either the apple will not be grasped with enough force and may slip, or the strawberry may be damaged by excessive force. Thus, different amounts of actuation for specific produce are necessary, realized through feedback from sensors to control the grasping force. One possible solution is the implementation of camera-based vision feedback for manipulation [13]. Additionally, cameras can be implemented to assist in detecting slippage from insufficient force [14]. The use of visuo-tactile sensing has also been used to detect the firmness of peaches [15]. However, the implementation of vision-based feedback is often highly dependent on the environment and requires often expensive equipment. Examples of these include a vision based tactile sensor for detecting features in soft produce such as strawberries [16] and peach firmness [17].

The utilization of sensing technology for robotics not only allows for the implementation of force and pressure feedback on grasping, but also for tasks suitable for the handling of produce. For example, they can be used for slippage detection to help determine the minimum amount of force required for grasping to avoid damaging the produce for both rigid grippers [18] and soft grippers [19]. Sensors can be used in combination with cameras for complex manipulation tasks [20]. Other uses include the estimation of the stiffness of the grasped object, such as a sensor used to detect the firmness of eggplants [21]. The data from these sensors can also be utilized to classify different types of fruits based on their distinct measurements [22]. Integrated sensors in the form of strain and pressure sensors can be implemented to assist in the process of categorizing items by their size and stiffness but can have lower accuracy due to the time delay of the sensor measurements [23]. Sensors utilizing other materials such as liquid metal with optional amplification layers can be utilized to increase the sensitivity of the pressure detection but require the use of potential hazardous material [24]. Additionally printable arrays of pressure sensors can be implemented with wireless communication for easier sensing of localized forces [25]. Similar pressure sensors are also made from stretchable and flexible materials that can allow for integration with soft grippers contorting in complex shapes; however, embedding the sensor inside the gripper can reduce its accuracy [26]. These previous sensors have the additional downside of having a relatively slow response, taking several to tens of seconds to reach a stable measurement. Faster sensors, such as a pressure sensor utilizing graphite, can be implemented with low-cost, quick fabrication with quick response time but with lower sensitivity and relative change in resistance [27].

In this work, a low-cost and quick-to-fabricate pressure sensor is implemented with robotic graspers for applications in produce handling. The timeframe of the sensor measurements is accelerated using an exponential curve fitting algorithm to predict the settled resistance while maintaining a high sensitivity in the measurements that can be used to distinguish different grasping forces and the stiffness of grasped objects. The design of the sensor is configurable in terms of shape, size, and spatial resolution to work with many applications and differently shaped robotic manipulators. The sensor is integrated with both a rigid gripper and a soft gripper, highlighting the capabilities to utilize it for different applications. The sensor was originally developed for detection of sea lamprey in aquatic environments [28], but the capabilities of the sensor to detect localized pressures in robust environments made it a suitable candidate for integration with robotic grippers. The sensor can be cheaply and easily fabricated using the piezoresistive material Velostat [29].

This work is organized as follows. In section 2, the sensor properties and fast estimation technique are discussed, along with the integration with a robotic grasper. Section 3 examines the relationship between the sensor measurements and the grasping force and material stiffness. Then in Section 4 several applications for produce handling are investigated. Finally, some concluding remarks as well as future work are given in Section 5.

2. Accelerated Estimation of Sensor Output

2.1 Sensor Description

The sensor utilized in this work is a carbon-infused piezoresistive pressure sensor. It features an array of flat pixels cut from a sheet of Velostat material connected through a set of electrodes cut from copper tape. The piezoresistive material and electrode are held in place with an outer substrate tape layer. A diagram of the general sensor design is shown in Figure 1a, which can be fabricated quickly (<15 minutes) for low raw material cost (<<1 USD). The copper tape and Velostat can be cut using a commercially available vinyl-cutting machine (Brother ScanNCut2 CM350) to produce a desired sensing pattern. After cutting, the copper is transferred from its original backing to the base of the sensor using transfer tape. The cut Velostat is then adhered to the sensor on top of the copper tape electrodes, and then the sensor is completed by adding a second base layer with additional copper tape electrodes. Finally, wires are soldered to copper tape electrodes to enable reading of the sensors from a measurement circuit. Further details on the fabrication of these sensors are detailed in [29].

When undergoing pressure, Velostat experiences changes in its resistive properties which can be measured using an external microcontroller. An increase in pressure to the sensor creates a relative decrease in resistance, while negative pressure increases resistance. This property is illustrated in Figure 1b. By comparing the measured resistance of individual pixels to a baseline resistance measured in the absence of pressure, an estimate of the pressure can be calculated. This enables not only an estimate of the overall grasping pressure for force estimation, but also to examine individual pixels with known locations on the sensor for further analysis of localized pressure. The design of the sensor is adjustable to the specific needs of the robotic gripper and application. The resolution of the pixels can be adjusted in terms of both the size, shape, and locations of the pixels themselves as well as the number of pixels. The two different designs of sensors utilized in this work are shown in Figure 1c.

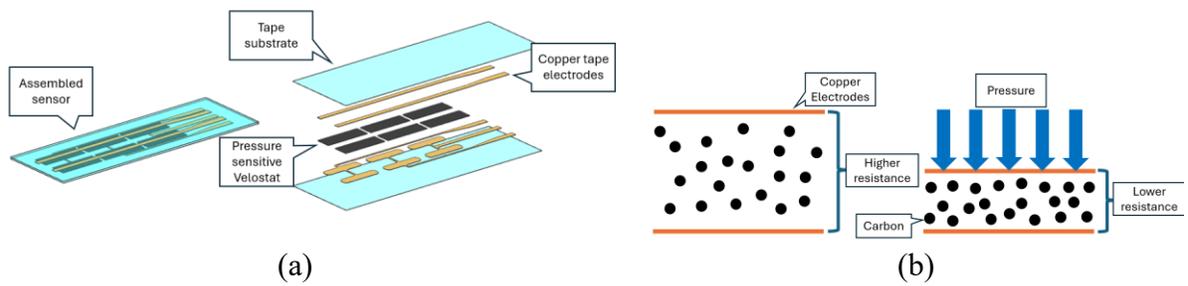

(a)        (b)

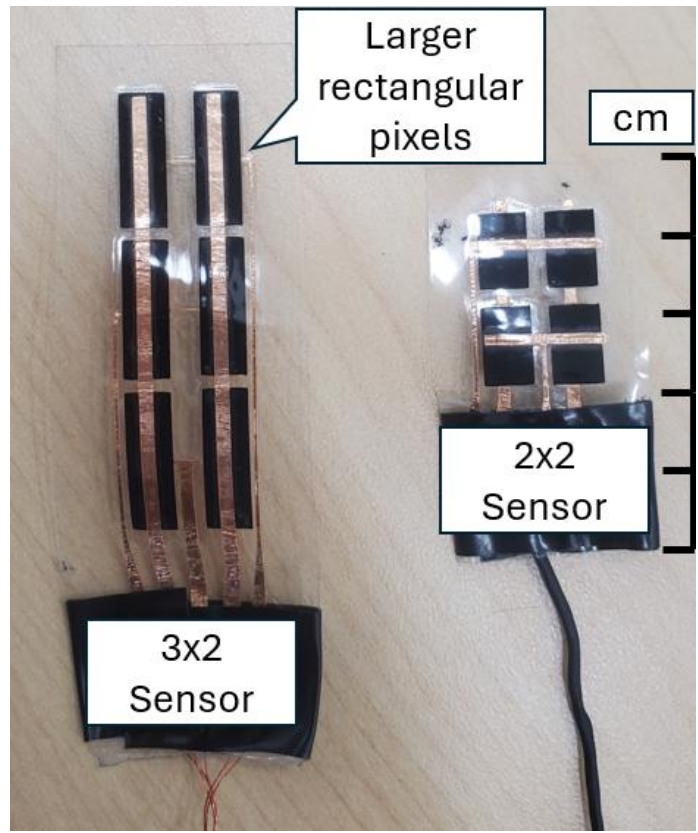

(c)

Figure 1: Physical properties of the sensor. (a) Diagram of the fabrication of the sensor, (b) Working principle behind piezoresistive behavior, (c) a 3x2 sensor with larger pixels and surface area coverage and a smaller 2x2 sensor with more precise localized pressure measurements.

Each individual pixel on the sensor gives a resistance measurement. The measurements of the pixels are fed through a set of two multiplexers that switch between the individual rows and columns of the sensor into an Arduino Mega 2560 analog-to-digital converter with 10-bit resolution. The sensor data is then fed over serial communication with a baud rate of 9600 to a laptop. This setup gives a sampling rate of approximately 15hz but could be increased with a higher baud rate. A diagram of the voltage divider is shown in Figure 2a, with the entire sensor setup shown in Figure 2b. In addition to the pressure readings, movement of the grippers causes a fast but large spike in resistance that leads to an exponential decay of the resistance measurements to a stable measurement. This can be seen both when the gripper opens and closes, indicating movement of the gripper as well as sudden shifts in pressure cause this effect. The measurements after a set waiting period are thus representative of the pressure. A plot of a typical measurement of the sensor is shown in Figure 2c when the robotic grasper grabs an apple, with both the closing and opening actuation of the robotic gripper noted by the actuation starts/stops indicator on the plot. It should be noted that based on the shape of the object grasped, not all pixels will make suitable contact with the object, leading to some pixels not decreasing in resistance as seen in the figure. Additionally, since each pixel has a unique resistance due to variance in the material and fabrication process, the data is converted instead into the relative resistance increase and decrease. A set of calibration data is used to normalize the individual pixels when under no external pressure with respect to a resting resistance $R_{avg}$. The individual measurements are then converted to a normalized form relative to these average measurements:

$$R_{rel}(t) = \left(\frac{R(t)}{R_{avg}} - 1\right) * 100\% \qquad (1)$$

In this work a reference voltage of 5V, and a resistor of 4.7kΩ are used in the voltage divider. To decrease the effects of noise and the variability of pressure on individual pixels from different grasps, in this work the average relative change in resistance across all pixels is calculated and used to derive specifics about the grasped object.

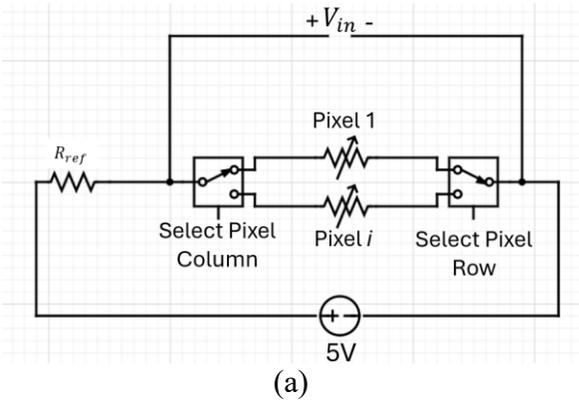
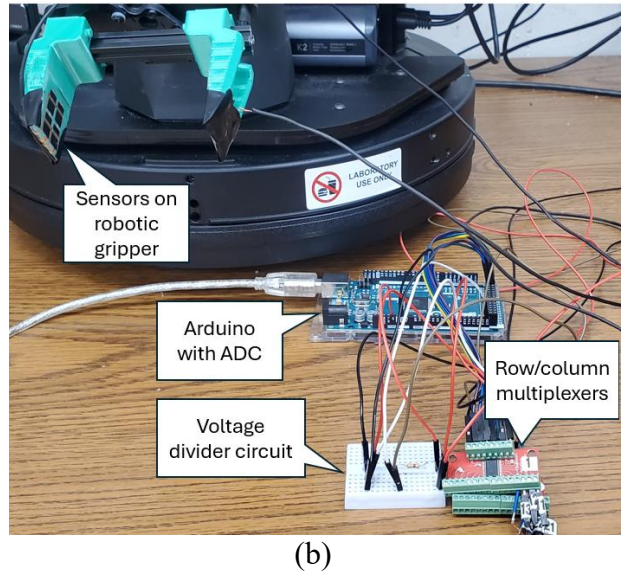

(a)      (b)

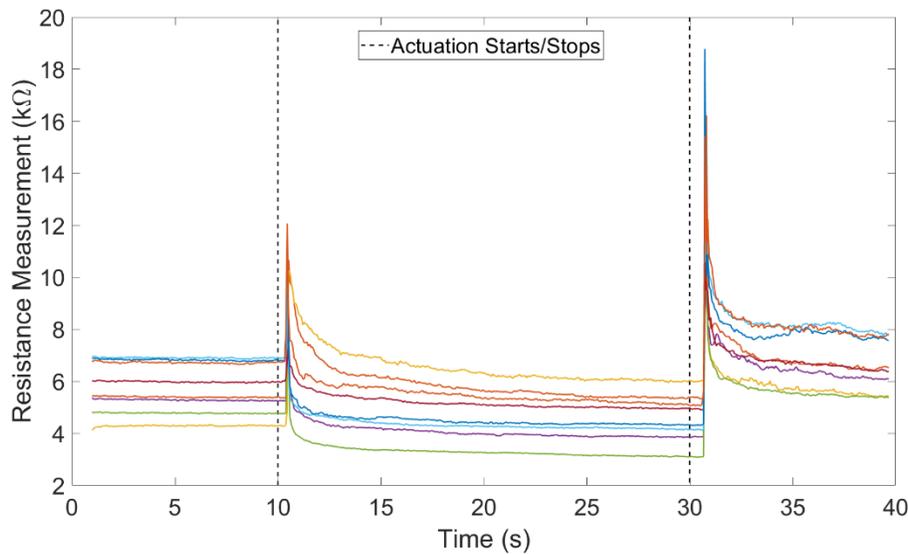

(c)

Figure 2: Arduino setup and measurements. (a) voltage divider circuit diagram, (b) physical setup to measure sensor with Arduino, (c) resistance measurement from the Arduino with no filtering.

The performance of the sensor under varying conditions is a crucial feature for its usability in produce handling applications where the conditions may vary greatly. The effects of temperature and humidity, as well as long term performance under repeated cycles are of concern. A set of experiments measuring the resistance over a period of 3 differing humidity and temperature conditions were conducted including one indoors at room temperature (21.9°C), and two outdoors in colder (14.8°C) and warmer (26.8°C) conditions. The results of these experiments are shown in Figure 3. Figure 3a shows the results in terms of absolute measured resistance, showing the difference in measured values as the conditions changed. The relative resistance measurements, shown in Figure 3b, can be used to reduce the variance in operating conditions to a smaller range.

An additional experiment of a long-term fatigue testing was done over 2,500 cycles with 10 seconds grasping and 10 seconds rest with a duration of over 13 hours. Results of these varying conditions on the sensor are shown in Figure 4. Figure 4a shows the absolute resistance measurements of the sensor as well as zoomed in sections during the earlier and later portion of the experiments. As seen in the plots, the sensor has a break-in period where the measurements are more sensitive before becoming more stable as time goes on. Figure 4b shows the same measurements but with the relative resistance in which the measurements were normalized at the beginning measurements of each cycle. This helps keep the average of the measurements centered around a fixed value over time and reduces the variance of measurements across cycles after the sensor goes through a break-in period.

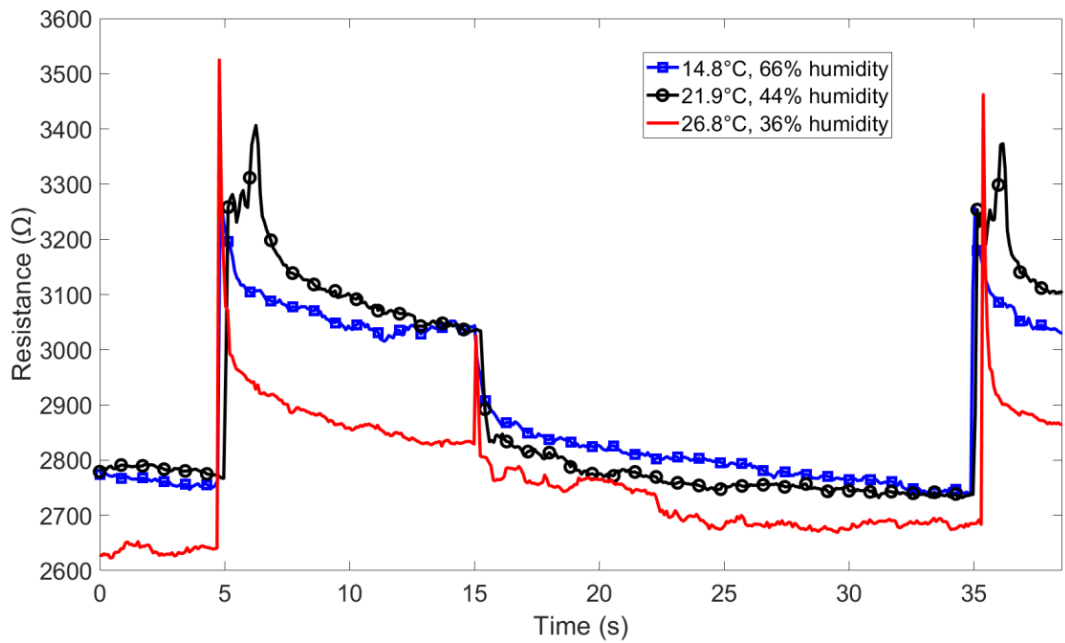

(a)

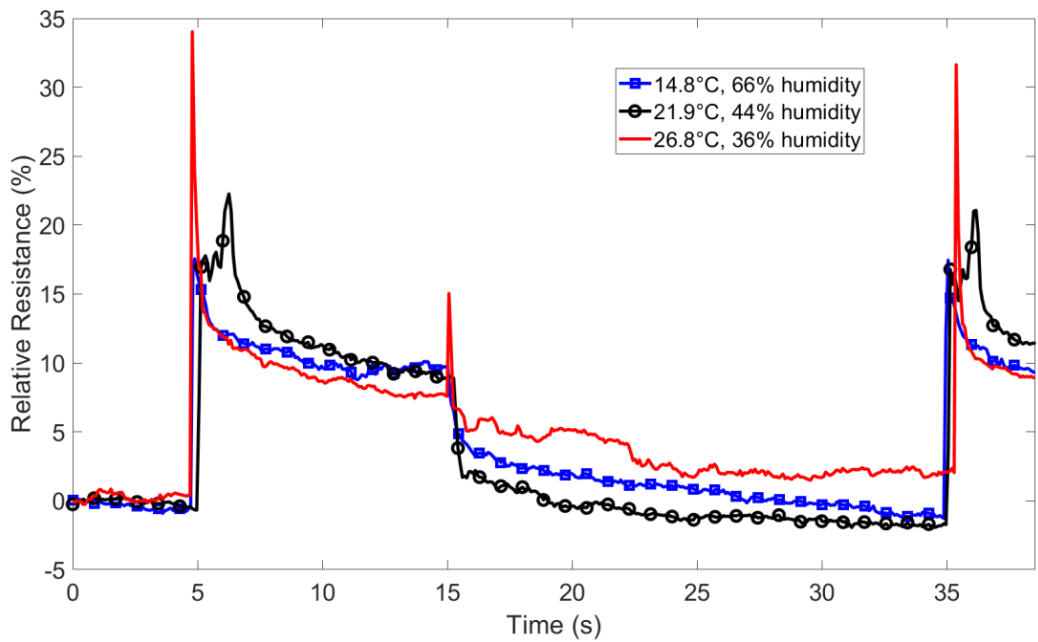

(b)

Figure 3: Sensor measurement under varying temperatures and humidity including one experiment inside at 21.9°C and two outside at 14.8°C and 26.8°C. (a) absolute resistance values recorded (b) relative resistance values.

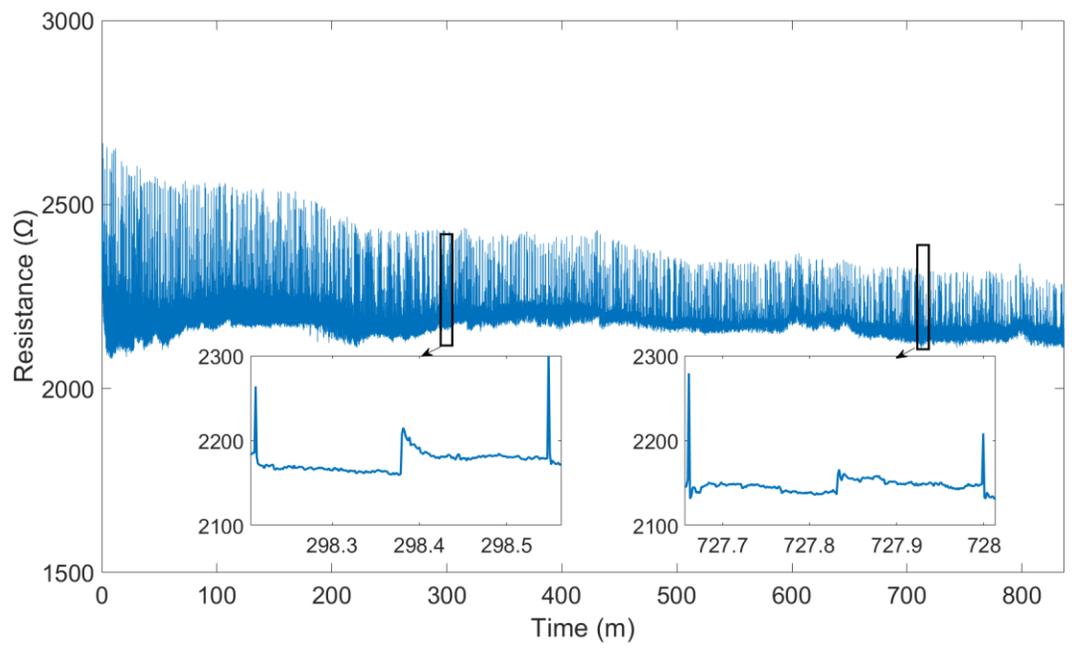

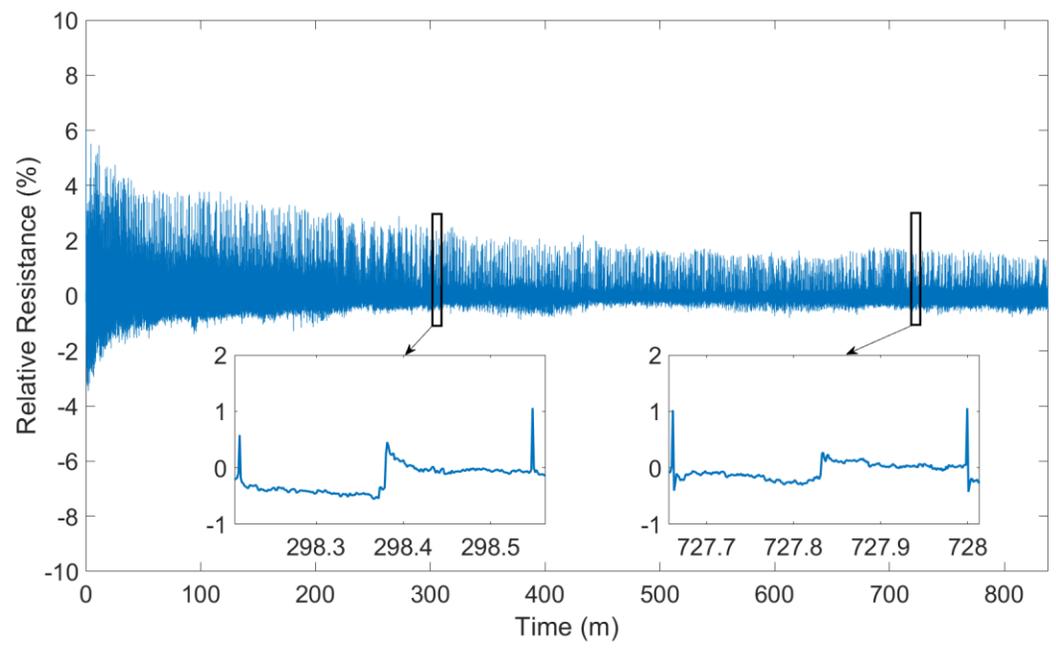

Figure 4: Sensor performance during long term use over 13 hours and 2,500 cycles. (a) absolute resistance values recorded (b) relative resistance values.

## 2.2 Exponential Curve Fitting

As shown in Figure 2c, when the sensor is in contact with an object, a large spike followed by slow settling of the resistances to a settled value is observed. This settled value is important for identifying characteristics of the object and grasping force. However, it is not always possible to wait extended periods of time for fully settled measurements. Even after a 10-second period, the sensor still experiences a small change over additional time. To reduce the time needed to estimate the steady state resistance, data in the transients are used to fit an exponential decay curve:

$$R_{est}(t) = A^* e^{-\lambda^* t} + C^* \tag{2}$$

$$A^*, \lambda^*, C^* = \underset{A, \lambda, C}{argmin:} \sum_{t=t_p+t_a}^{t_c} \left\| R_{rel}(t) - Ae^{-\lambda(t-t_p)} + C \right\|_2$$

$$t_p + t_a < t_c \tag{3}$$

where $A^*$ corresponds to the initial spike in resistance, $\lambda^*$ is the decay factor, and $C^*$ is the settled resistance which can be used as an estimate for the actual resistance value after an infinite amount of time. In equation (3), three points in time are used:

1. $t_c$, a cutoff time for data to be collected after actuation
2. $t_p$, the time of the peak resistance within the cutoff time occurring at the time of initial contact
3. $t_a$, a period of time after the peak to wait for less-noisy measurements

The value of $t_a$ was heuristically chosen to be 0.5 seconds, as this removes some of the initial data after the peak where a faster time scale is dominant to improve the curve fitting results. To determine an appropriate amount of time to allow for the approximation of the curve, a set of data from four different silicone sheets (described in Section 3.1) and five different force levels, varied by different grasping widths of the gripper on the silicone sheets for each material were collected. These grasped were repeated a total of 10 times for each material at each grasping width for a total of 200 datasets. Different periods of time were selected as a cutoff period for the exponential curve fitting, acting as the total delay between actuation and sensor estimation, and compared to the assumed settled resistance after 20s, the total duration the gripper was closed, to calculate the error. While this method reduces the time to estimate the settled resistance values, it still creates a time-delay, making the sensor more suitable for applications with stable grasping conditions over several seconds and an instantaneous estimate is not needed. These results are compared to the error from the recorded resistance at the same cutoff time. These results are shown in Table 1 and Figure 6. From these experiments, a cutoff time of 2.5s was selected as a compromise between high accuracy and faster estimation. However, variations in the actuation speed across grippers would likely alter initial resistance measurements, possibly necessitating different measurement delays. Figure 5 shows the median error results from fitting an exponential curve to a set of data from the sensor with a cutoff time of 2.5s.

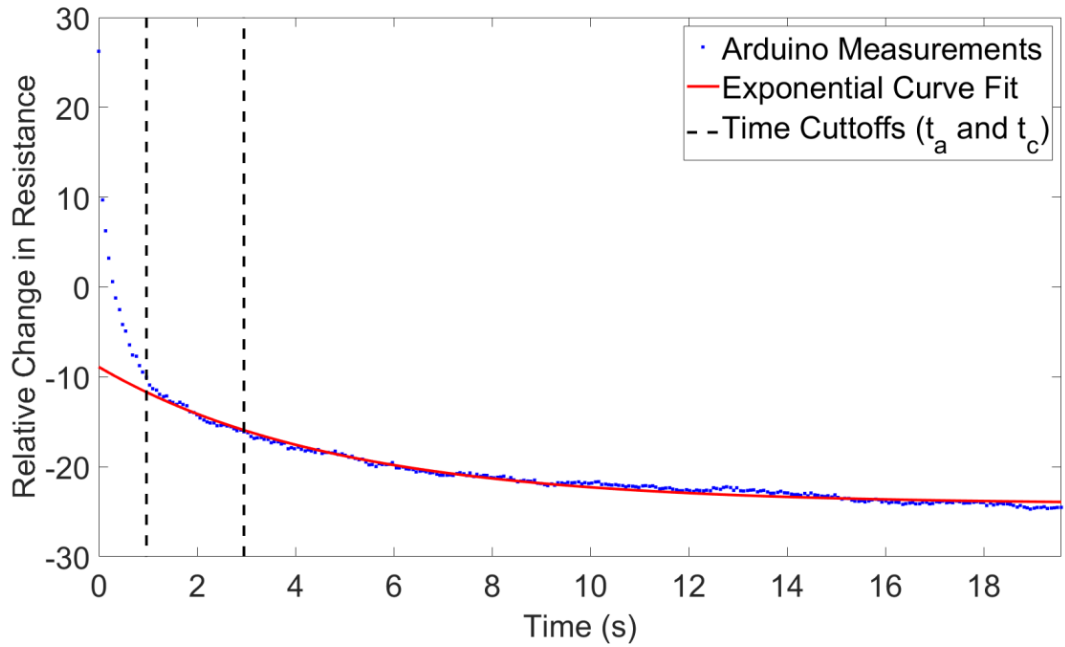

Figure 5: Exponential curve fit of median error from data between 0.5-2.5 seconds.

Table 1: Comparison of errors with settled relative resistance using exponential decay curve fitting and the recorded resistance value at the same cutoff time.

| Cutoff time ($t_c$) | 1s | 2s | 2.5s | 3s | 4s | 5s | 6s | 7s | 8s | 9s | 10s | 15s | 20s |
|---|---|---|---|---|---|---|---|---|---|---|---|---|---|
| Exponential decay resistance error (%) | 12 | 2.9 | 2.5 | 2.3 | 2.6 | 2.8 | 2.7 | 2.7 | 2.5 | 2.3 | 2.0 | 1.0 | 0.58 |
| Recorded resistance error (%) | 15 | 10 | 9.2 | 8.2 | 6.6 | 5.4 | 4.6 | 3.9 | 3.2 | 2.8 | 2.4 | 0.86 | "0" |
| Error reduction ratio (%) | 17 | 72 | 72 | 72 | 60 | 49 | 40 | 31 | 24 | 20 | 15 | -22 | N/A |

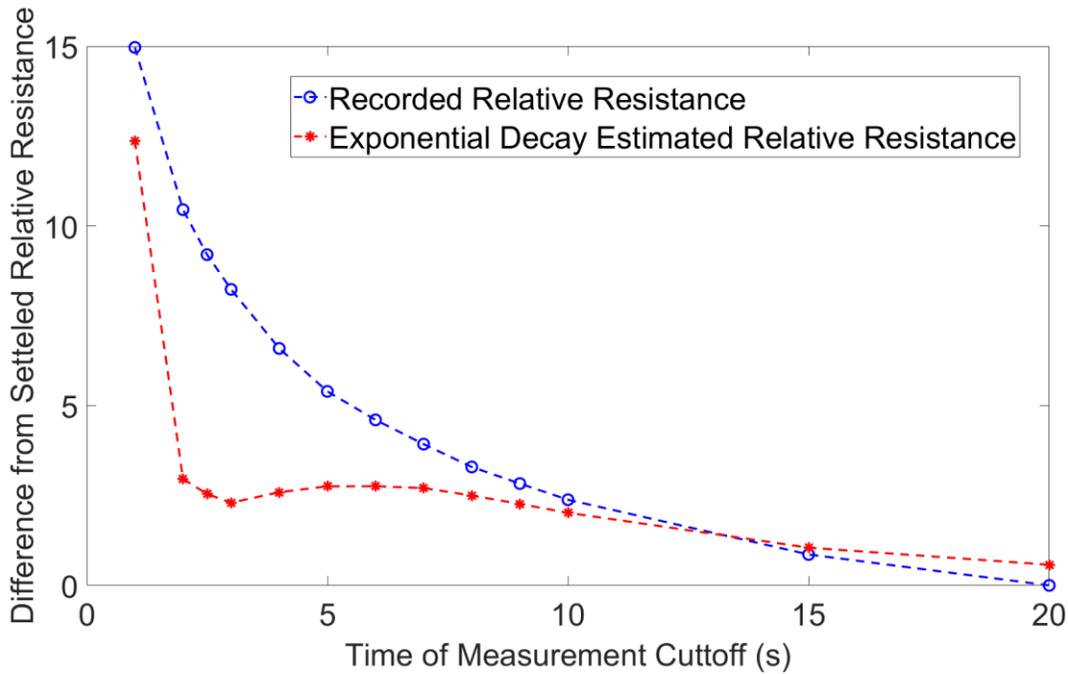

Figure 6: Plot of errors compared to the settled relative resistance from using exponential curve fitting with certain cutoff times and using the recorded resistance at the same time.

3. Estimation of Stiffness and Force

3.1 Robotic Gripper Integration

The sensor described in Section 2.1 was integrated with a set of rigid robotic fingers. These fingers are incorporated with the Locobot WX250S mobile robotic platform. This platform carries a 6 degrees of freedom robotic manipulator with an end-of-arm tooling including the robotic fingers. These fingers are controlled via a servo motor and can be set to a variety of different modes including a width-controlled mode used for this work. In this mode, the distance between the two fingers is controlled, and the robot attempts to turn the servo motor to achieve this width regardless of the force required to do so. The servo motor is set to lock if a certain torque is exceeded, so precise monitoring of the output force is important. The sensors were attached to the pair of grasping fingers with each containing an array of 2x2 pixels, for a total of 8 pressure readings on the fingers. The sensor is pictured on the right of Figure 1c, and an image of the sensor attached to the robotic gripper is shown in Figure 7.

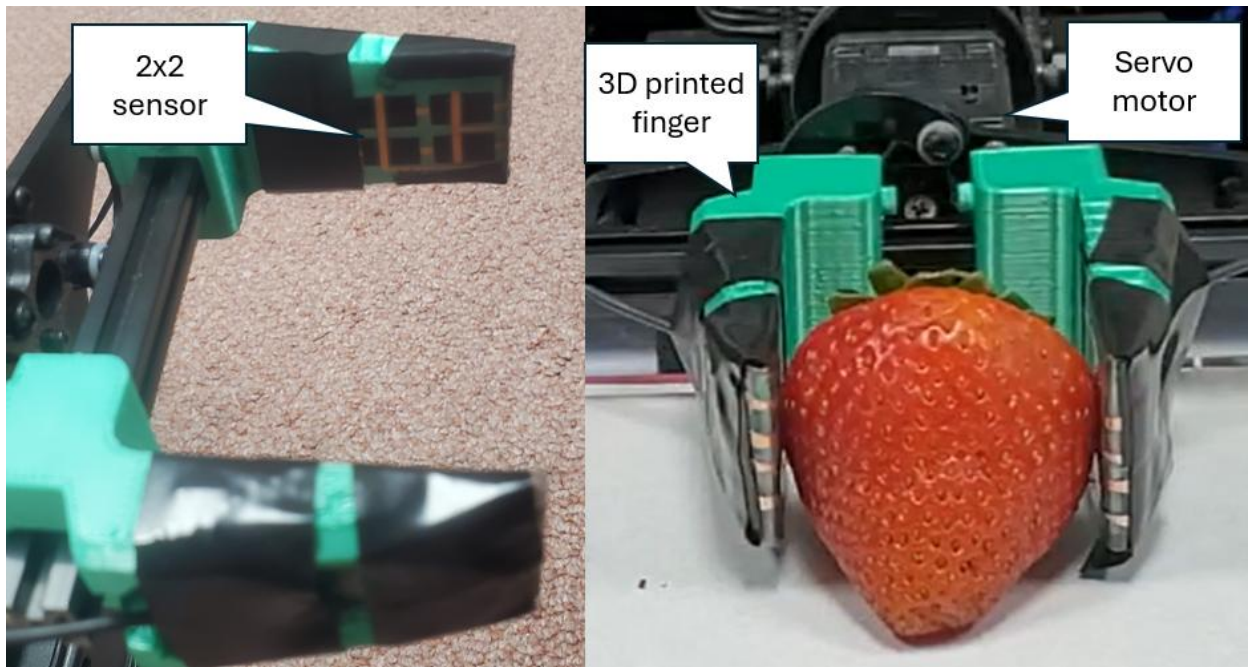

Figure 7: Rigid gripper with sensor

Supplementary video 1 demonstrates a simple use case of the sensor with the Locobot and rigid robotic gripper. In this video, thresholds are used to identify if the sensor on individual fingers of the robotic gripper is in contact with an object. This can be used to detect human input for manual control, to confirm that an object was successfully grasped, and for identification that the object has been removed or slipped, all of which are shown in the video.

3.2 Stiffness Estimation

The stiffness or hardness of an object is essential to know for proper grasping, especially for grippers with positional control such as the rigid gripper used in this work. A soft object will deform more than a stiffer object under the same force, but if equal grasping force is necessary to lift both objects, then a different amount of deformation is necessary to grasp both. This can make grasping of an object of unknown stiffness, even with a known size, difficult to perform with positional control. Produce such as apples, avocados, and bananas can have a wide variety of stiffnesses as they ripen, making them a prime candidate for stiffness estimations. By examining the measurements of the sensor, the stiffness of a variety of materials can be estimated.

The pixels on the sensors provide measurements of the individual pressure induced on them. For more rigid objects, a higher pressure is created under the same displacement causing pixels with contact to the object to have a higher change in resistance measurements under the same actuation. By examining these measurements, the overall stiffness of the object can be estimated and used to ensure proper grasping force without damaging the produce. Additionally, with both the size and stiffness estimates, it is possible to categorize many different types of produce.

To isolate the capability of the sensor to detect differences in stiffness, a set of materials with known stiffnesses need to be utilized. To perform these tests, different silicone pads were created with casting, where DragonSkin 30, DragonSkin 20, DragonSkin 10, and EcoFlex 10, with their respective shore hardness values of decreasing stiffness as A-30, A-20, A-10, and 00-10 were used. These pads were grasped a total of 10 times across trials with a container made to hold them in place for grasping with the rigid gripper. The experimental setup is shown in Figure 8a. With this setup, the differences in measurements could be attributed to only the differences in stiffness and not size or shape. As expected, the stiffer silicone pads created a larger decrease in resistance. The average pixel measurement of the three different pads is shown in Figure 8b with the values reported in Table 2.

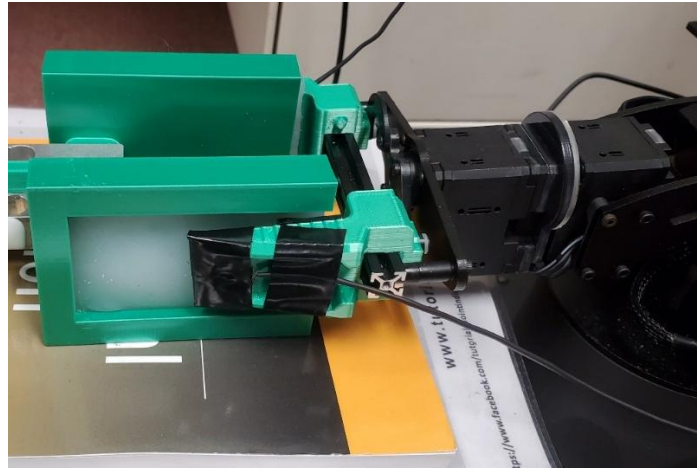

(a)

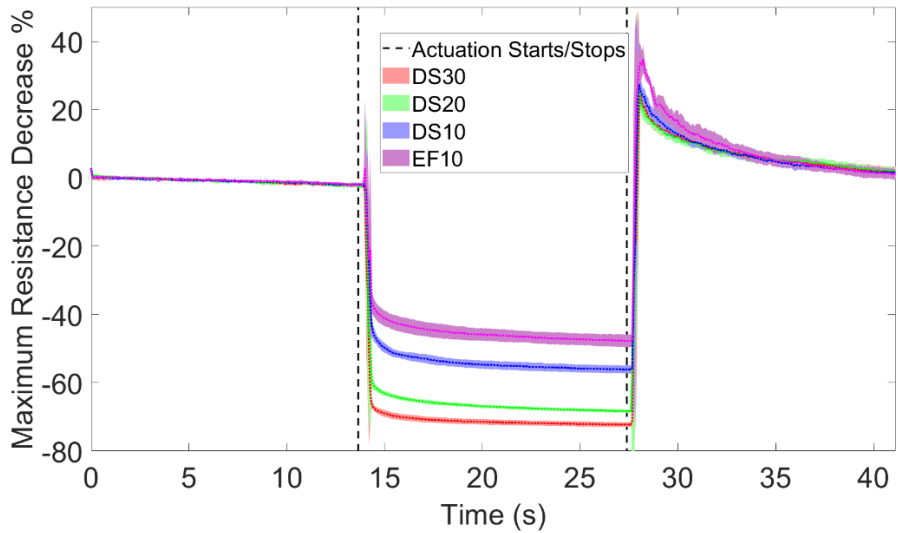

(b)

Figure 8: Silicone sheets experimental setup and results. (a) physical setup of robotic grippers with sensors and silicone sheets, (b) comparisons of relative resistance from four different silicone sheets with mean values over 10 experiments in dashed line with ±1 standard deviation shown by the shaded regions.

Table 2: Silicone sheet stiffness data

| Material | Ecoflex 10 | Dragonskin 10 | Dragonskin 20 | Dragonskin 30 |
| --- | --- | --- | --- | --- |
| Shore hardness | 00-10 | A-10 | A-20 | A-30 |
| Relative change in resistance | -46.0% ± 1.8% | -54.7% ± 1.2% | -66.9% ± 0.52% | -71.5% ± 0.74% |

In addition to the silicone pads, a selection of 6 exercise balls were utilized to measure different stiffnesses. These exercise balls are rated in terms of grip strength, ranging from 33lbs to 66lbs, with the larger grip strength corresponding to a higher stiffness. These balls were tested with the rigid gripper over a set of 30 trials as shown in Figure 9a. The data from three of these exercise balls is shown in Figure 9b with the numbers included in Table 3. As expected, as stiffness increases, the resistance also decreases showing the capability of the sensor to detect differences in stiffness values. Additionally, when examining the pixel with the maximum decrease in stiffness, a linear correlation between this value and the grip strength of the balls is present with an $R^2$ value of 0.92. This relationship is shown in Figure 9c.

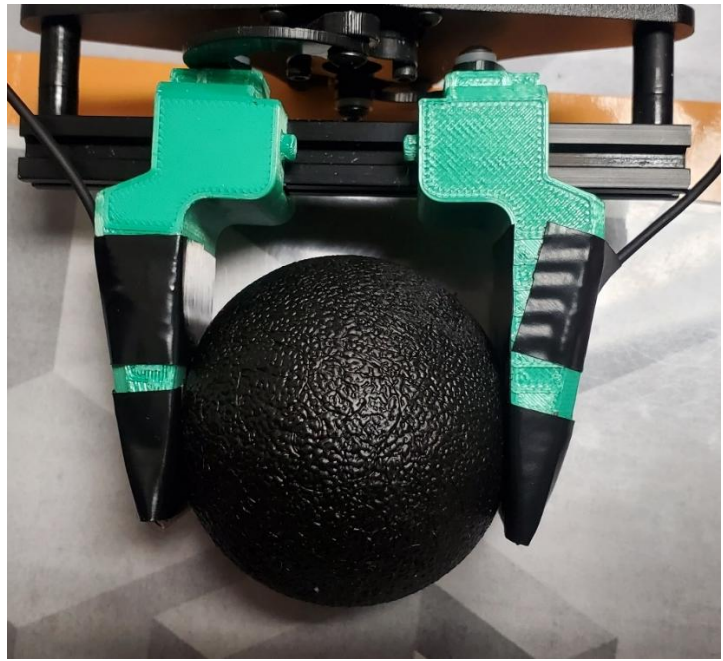

(a)

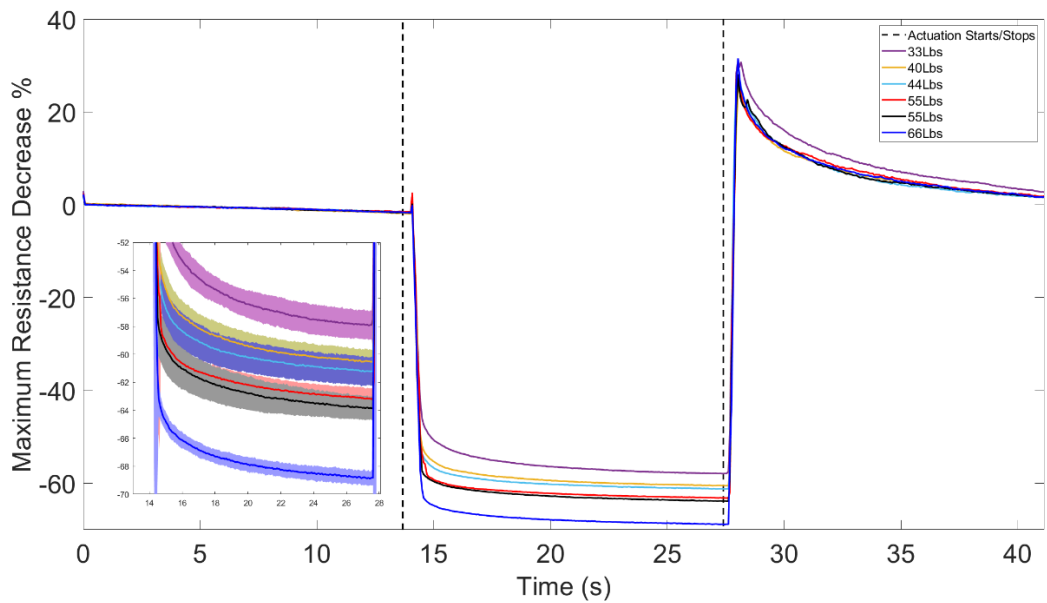

(b)

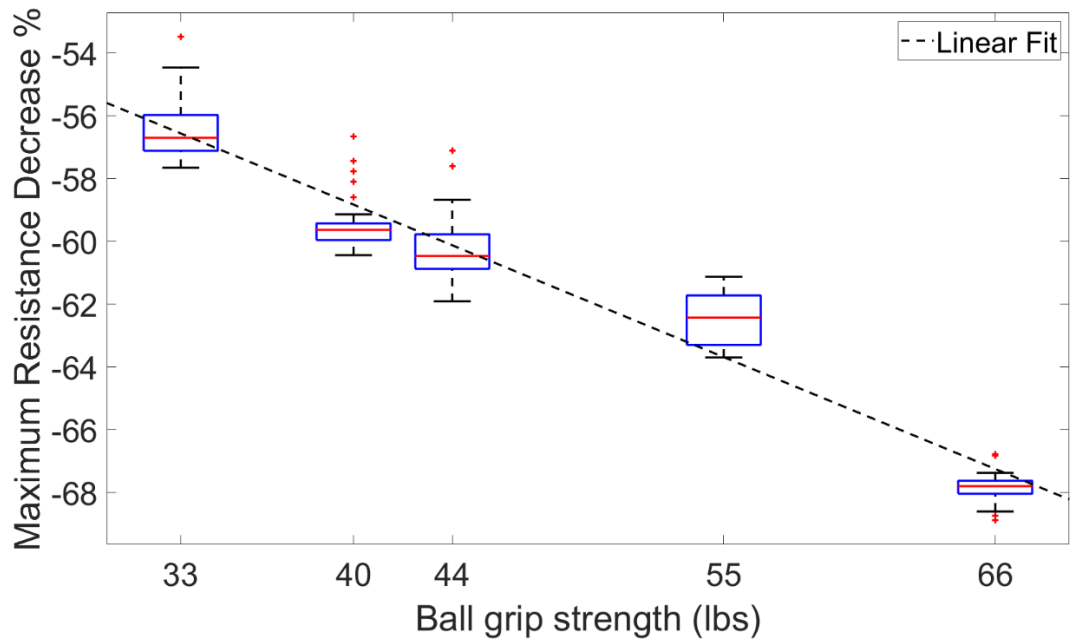

(c)

Figure 9: Exercise ball experimental setup and results. (a) Physical setup of robotic gripper with sensor grasping exercise ball, (b) plots of mean relative resistance from different balls including shaded regions of ±1 standard deviation across the 30 trials in the zoomed in section, (c) linear fit of the measured resistance and advertised grip strength of balls.

Table 3: Exercise ball relative resistance measurement

| Grip strength | 33lbs | 40lbs | 44lbs | 55lbs | 55lbs | 66lbs |
|---|---|---|---|---|---|---|
| Relative change in resistance | -56.4% ± 0.98% | -59.4% ± 0.95% | 60.2% ± 1.1% | -62.2% ± 0.76% | -62.8% ± 1.14% | -67.9% ± 0.49% |

Finally, a third set of stiffness measurements using a set of compression springs with known stiffness values were used to evaluate the relationship between the sensor measurement and stiffness of the material. A set of parallel plates with varying amounts of springs were used to create a setup with a customizable stiffness and grasped at the same width with the gripper. A set of four different setups of springs were used to collect samples across varying stiffness, with 10 grasps being performed on each setup. The hardware setup for these experiments is shown in Figure 10a with the resulting relationship between the change in resistance and stiffness shown in Figure 10b. The data shows a strong linear correlation between the sensor measurements and the stiffness of the springs with an $R^2$ value of 0.96.

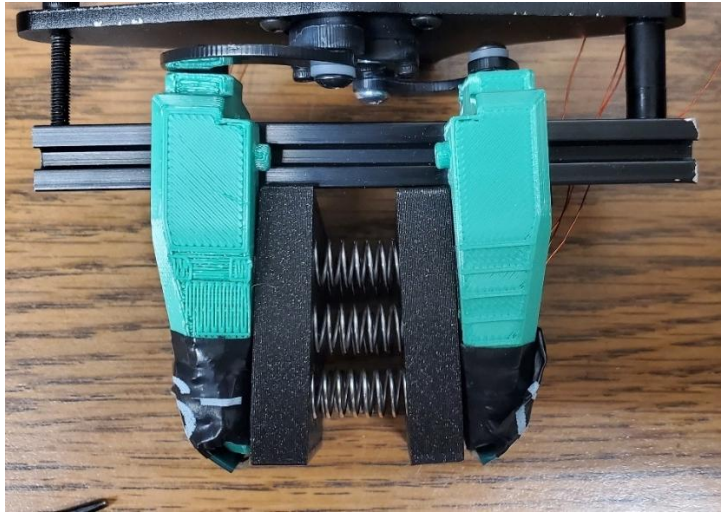

(a)

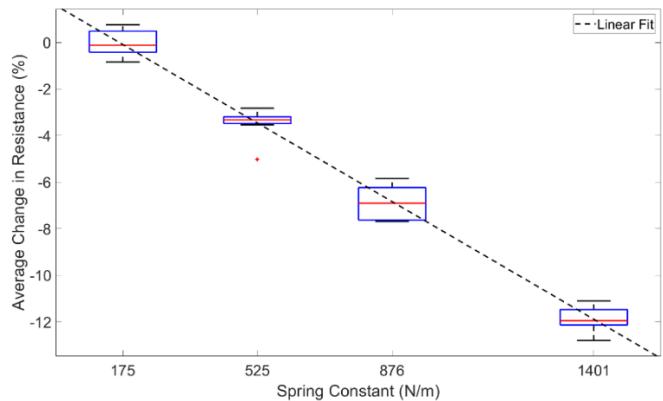

(b)

Figure 10: (a) Spring experiment hardware setup using parallel plates and variable number of springs. (b) Results of the resistance measurement vs the spring constant across 4 experiments over 10 trials each.

3.3 Force estimation

A crucial feature in robotic grasping is controlling the force and pressure at which the grippers are outputting to the object. With too little force, the object will not be properly grasped and may slip from the grippers. With too much force, the object could be damaged by the grippers. By using the sensors in this work, the output force of a gripper can be estimated. A higher force on the object will result in the pixels reducing in resistance. If the sensor is calibrated properly, the resistance measurements can be used as feedback to control the pressure to a desired value.

Experiments were done utilizing a pair of digital load cells (ShangHJ 1kg, HX711) compressed by the rigid robotic gripper shown in Figure 11a and 11b. These were calibrated using a set of known weights with a sensitivity of 1g and mounted with custom 3D printed plates to hold the silicone pads as well as keep them in place along the axis of the robotic gripper attached around the T-slot bar. The load cells included the four silicone pads of differing stiffnesses described in Section 3.2.1 that the robotic fingers press on to induce a grasping force. The fingers were moved to a set of tighter grasping widths to slowly increase the force applied while observing the sensor measurements. Each grasping width was repeated 10 times for a total of 50 trials. The collected force and sensor measurements are shown in Figure 11c for the four different silicone pads.

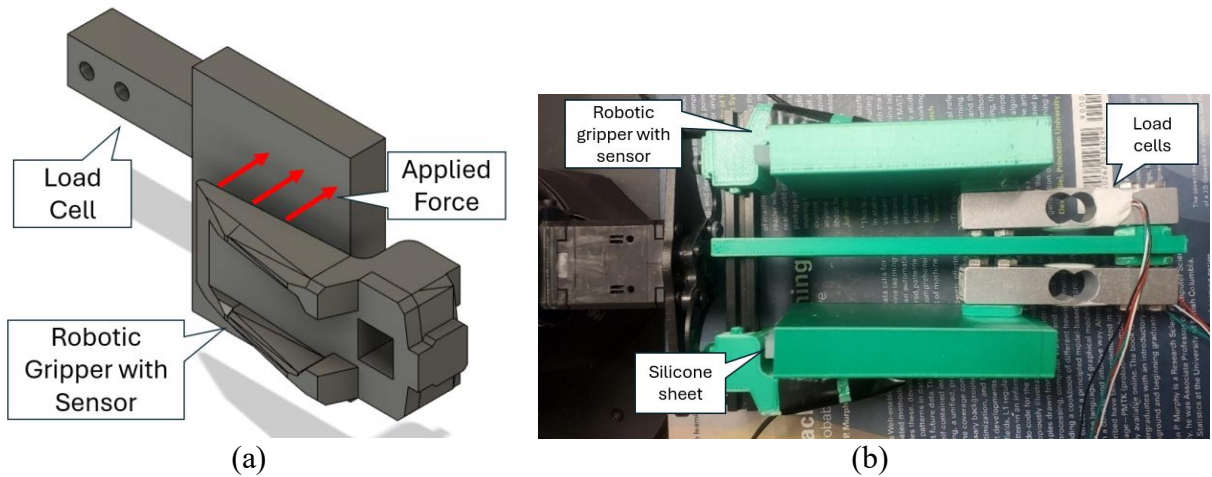

(a)    (b)

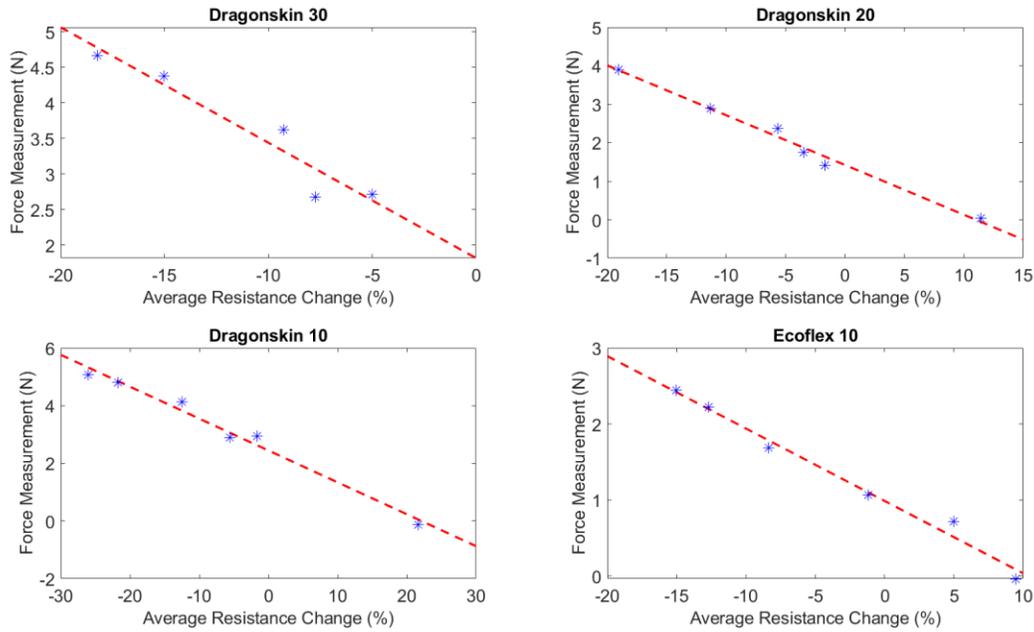

(c)

Figure 11: Load cell setup and results. (a) CAD diagram of load cell layout, (b) physical setup of robotic grippers with sensors and load cell, (c) linear fit between output force and resistance change between four different silicone sheets.

From the collected data, there is a strong linear correlation in the output force to the measured resistance change of the sensor. Based on this, it would be possible to estimate the current grasping force, as well as use this data to control the force to a desired amount. The properties of the associated fitted linear approximations are shown in Table 4. The slope of each fitted model represents the increase in force with respect to the percentage decrease in resistance. From the data, it can be observed that an increase in the silicone sheet stiffness is associated with a steeper slope, that is to say the resistance measurements of the sensor decreases more per applied unit of force.

Table 4: Properties of force approximation models from four silicone sheets

| Silicone Sheet | $R^2$ | Slope (N/%) | Y-intercept |
|---|---|---|---|
| Dragonskin 30 | 0.917 | -0.163 | 1.81 |
| Dragonskin 20 | 0.986 | -0.129 | 1.42 |
| Dragonskin 10 | 0.983 | -0.111 | 2.45 |
| Ecoflex 10 | 0.985 | -0.0953 | 0.987 |

The exponential curve fitting technique was utilized to estimate the force using the parameters from Table 4 and compared to utilizing the recorded resistance at 2.5s, 10s, and 20s. The results of these estimations are shown in Figure 12. From the results, it can be seen that the exponential curve fitting provides a significantly better estimation of the force than taking the raw sensor measurements at both 2.5s and 10s. While not being as accurate as waiting a full 20s for a settled resistance value, it does quickly provide a reasonable estimation when time is an important factor.

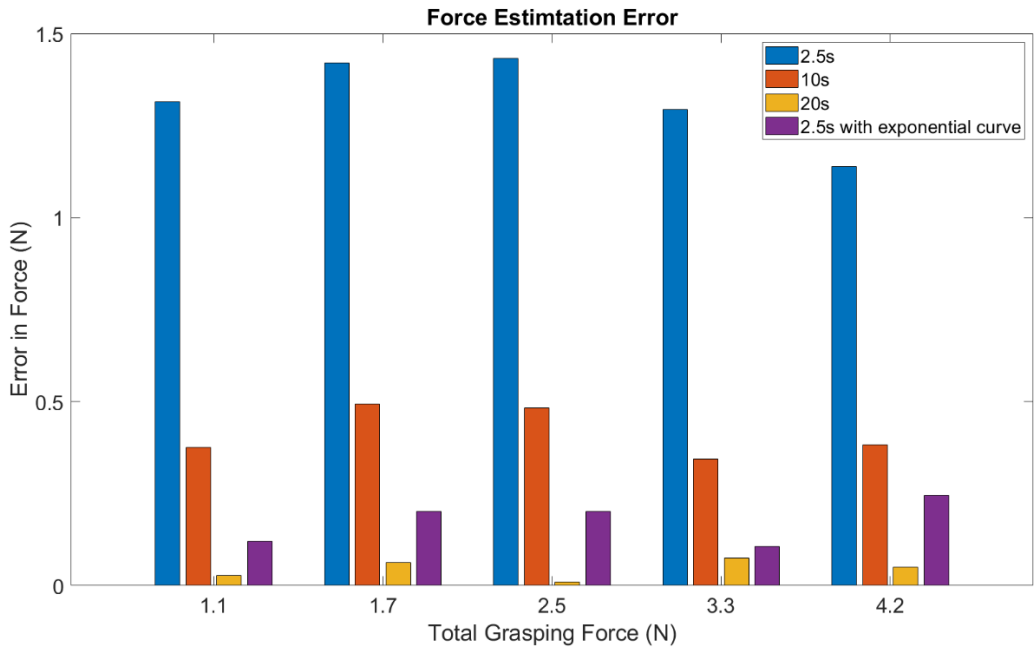

Figure 12: Force estimation using recorded resistance at various times and an exponential curve fit at 2.5s.

Table 5: Force estimation errors using recorded resistance at various times and exponential curve fit at 2.5s.

| Technique used | 2.5s | 10s | 20s | Exponential |
|---|---|---|---|---|
| Average error | 1.32 ± 0.12N | 0.42 ± 0.068N | 0.045 ± 0.026N | 0.14 ± 0.12N |
| Percent error | 64.4% | 20.0% | 2.07% | 7.94% |

## 4. Produce Handling Applications

### 4.1 Size Estimation

By utilizing the sensor, key characteristics of the object can be identified without prior knowledge. One of these parameters is the size of the object. The sensor is capable of identifying when the gripper comes into contact with the object. For the rigid gripper, the gripper can be incrementally closed at certain resolutions until contact is made with the object. Since the gripper needs to slightly squeeze and not just make contact with the object, the previous finger-width increment can be used as an estimate of the size of the object. Figure 13 shows the process used to estimate the size of an unknown object. From this point, an additional amount of width actuation can be used to close the gripper to fully grasp the object. This allows for the gripper to securely grasp an object of unknown size without risking damaging the object with excessive force. Supplementary video 2 shows this process of size estimation being implemented with three objects of different sizes, where the gripper closes until it verifies sufficient pressure is generated to lift the objects without damaging them. Additionally, supplementary video 3 shows the use of the gripper in identifying if there is an object present at all, opening and closing until something is present.

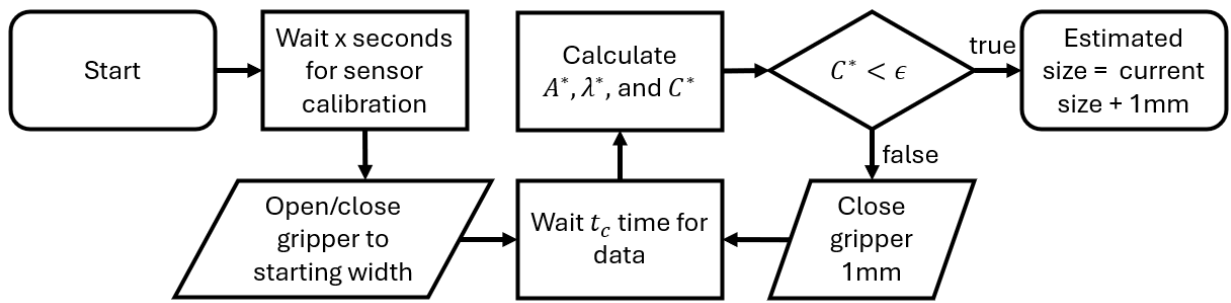

Figure 13: Flowchart of size estimation algorithm used for strawberry size estimation.

This technique can be used to handle produce that can vary in size such as strawberries. Experiments were performed with the rigid gripper using the process described above by closing in increments of 1 mm to estimate the size of 10 different strawberries, with the size being described as the diameter of the strawberry along the axis grasped at. For this experiment, an $\epsilon$ value of -10% was used to determine if the gripper contacted the strawberry. The strawberries varied in size from 31-41 mm as measured with a caliper to the nearest mm. The estimated sizes versus the measured size are shown below in Figure 14. The size estimation had an RMS error of 0.89 mm, with an average percent error of 2.2% with respect to the size of the strawberry with all measurements falling within ±1 mm of the true size.

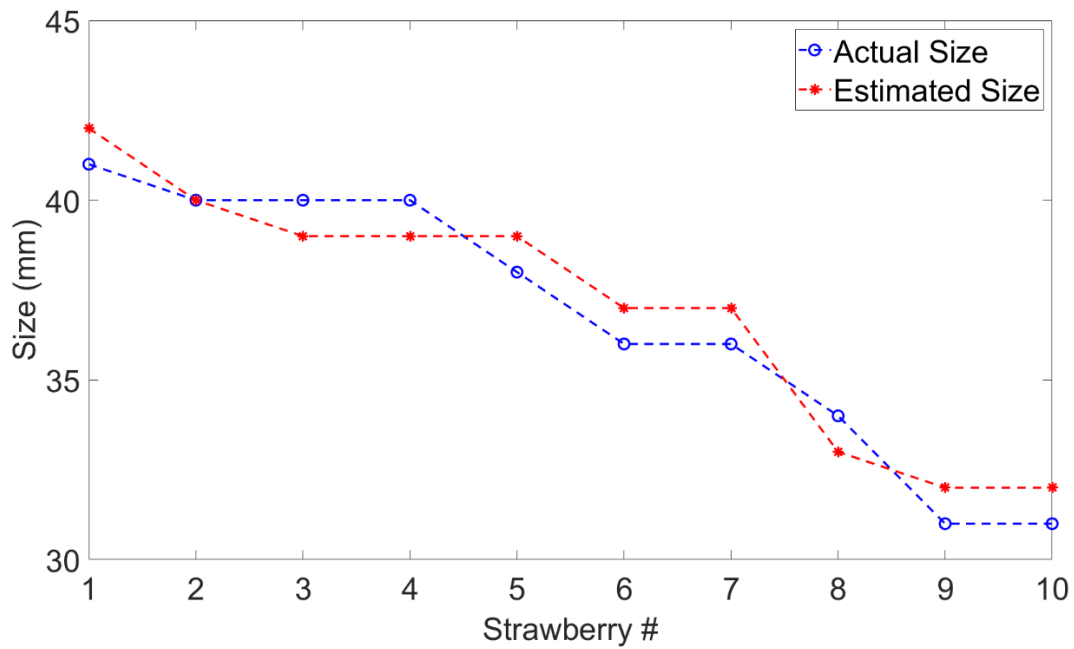

Figure 14: Strawberry size estimation and actual size.

4.2 Ripeness Monitoring

One possible use case for the sensor could be in identifying the ripeness of certain produce. If the produce changes in physical properties such as stiffness as it ripens, the sensor is capable of detecting this change as shown in the previous section. This could be utilized for cases such as identifying if produce is in a condition to be harvested from its source, or in terms of distribution in selecting items of certain ripeness to deliver them at their ideal state. For these experiments, in addition to the rigid robotic gripper, a pair of sensors was also implemented with a set of soft robotic fingers to monitor the ripeness of an avocado. The use of a second set of robotic grippers with the sensor here is used to demonstrate the adaptability of the sensor to easily be implemented with a variety of hardware. The soft fingers are a pair of pneumatically actuated graspers fabricated with the silicone material DragonSkin 30. When pressurized air is sent to the fingers, a bending motion is induced, which will cause the two fingers to push on and grasp the object in between them. The range of working air pressure for the grippers is a maximum of 50 kPa. As opposed to the rigid grippers which utilize position control, the control of the soft fingers is more directly tied to the force output. While the inherent softness of the fingers makes them less likely to damage produce, it can be more difficult to ensure that they are maintaining proper contact and force on the object for proper manipulation. The use of this pressure sensor can enable feedback to create sufficient grasping of the object. Both soft fingers were augmented with a 3x2 sensor array for a total of 12 pressure readings on the fingers. The sensor is shown on the left of Figure 1c, and the sensor attached to the soft grippers are shown in Figure 15. Supplementary video 4 shows the process of gathering data on an avocado using the soft fingers.

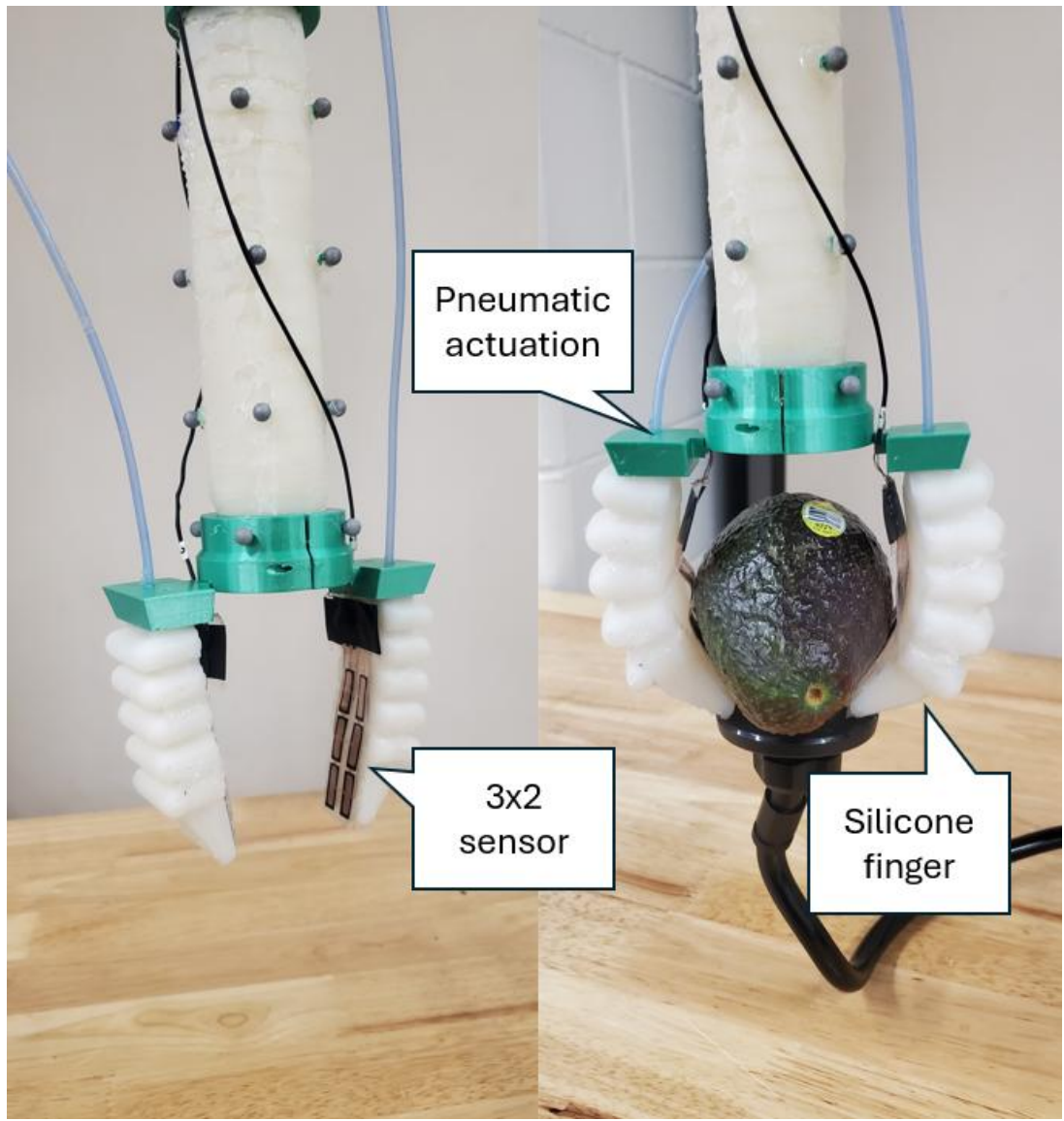

Figure 15: Soft gripper setup.

To test the feasibility of this process, an avocado was tested over several days to monitor ripening. Images of the Avocado over the final four days are shown in Figure 16a as it ripens. Avocados are known for the quickness that they ripen at, as well as the large difference in stiffness as they ripen. The average change in resistance taken over a set of 10 trials each day of the avocado from the rigid gripper over the tested days are shown in Figure 16b, with the soft gripper also producing a similar trend shown in Figure 16c. With the soft gripper, the avocado reached full ripeness a day earlier due to being more ripe at purchase. Utilizing this information, it would be possible to identify what stage of ripening an avocado is at, and the use case could expand to other fruits if similar experiments were performed.

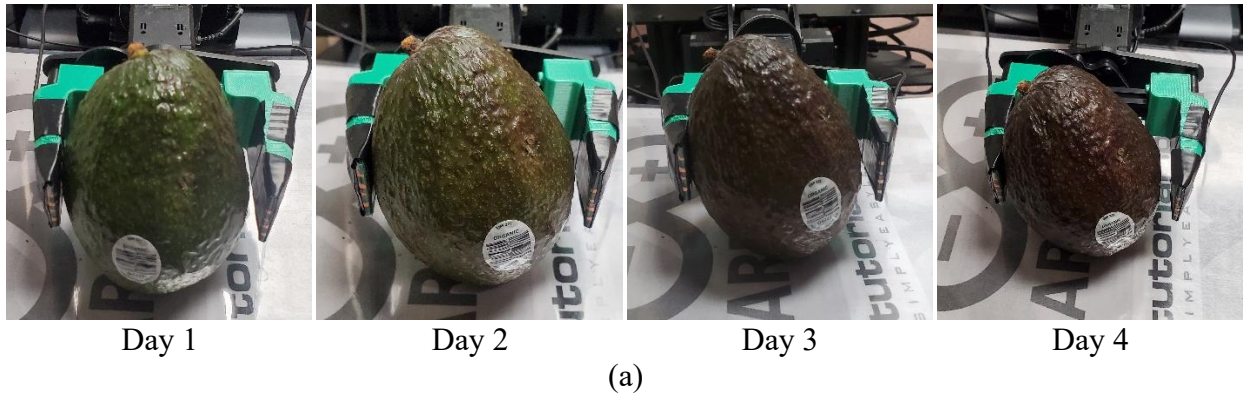
Day 1　　　　　Day 2　　　　　Day 3　　　　　Day 4

(a)

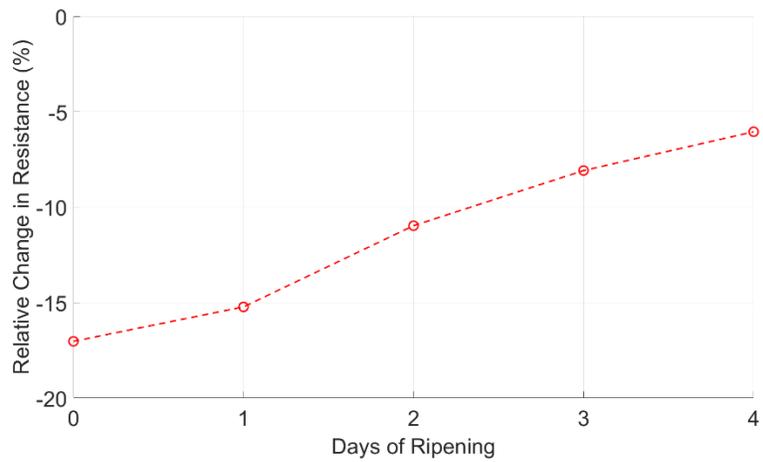

(b)

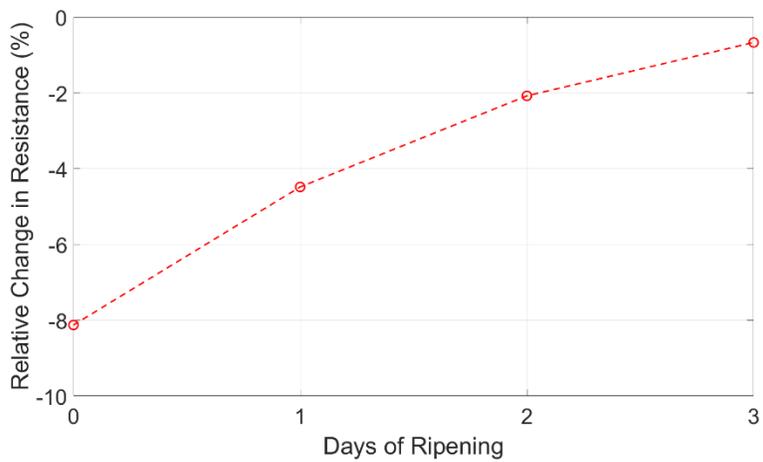

(c)

Figure 16: Avocado ripeness monitoring and results. (a) Avocado over 4 days of ripening, (b) data collected using rigid gripper, (c) data collected using soft gripper.

## 4.3 Bruising Detection

Another possible application of the sensor would be for its use case in identifying possible damage to produce. This could come in the form of diseases or bruising that alters the stiffness of the item. For example, apples, when bruised, create a soft spot that could be identified by the sensor. To validate this possibility, sensor measurements of the same apple were taken before and after intentionally bruising one side of the apple by dropping it onto a cement floor, repeated over 10 trials as shown in Figure 17a. This created a soft spot on the apple that can be identified with the sensor producing a lower change in resistance with an identical grasping width. A plot showcasing the difference in sensor measurements between the bruised and unbruised apple is shown in Figure 17b with the dashed lines representing the mean resistance values and the shaded regions showing ± 1 standard deviation across the 10 trials. The data displays an increase in resistance measurement after the apple was damaged due to the softening of the bruised area. Additionally, an exponential curve can be fit to the data as described in section 2.2 to quickly categorize the produce as damaged or not, as even after 15 seconds the measured resistance still had not settled. By examining the resistance measurements when grasping, it would be possible to identify anomalous stiffness values in different produce. This could enable the detection of bruises during distribution without increasing the process time. The measured changes in stiffness before and after bruising, as well as examining specifically the measurements on the pixels contacting the bruised area, from both the measured resistance and the estimated resistance using exponential curve fitting are shown in Table 6.

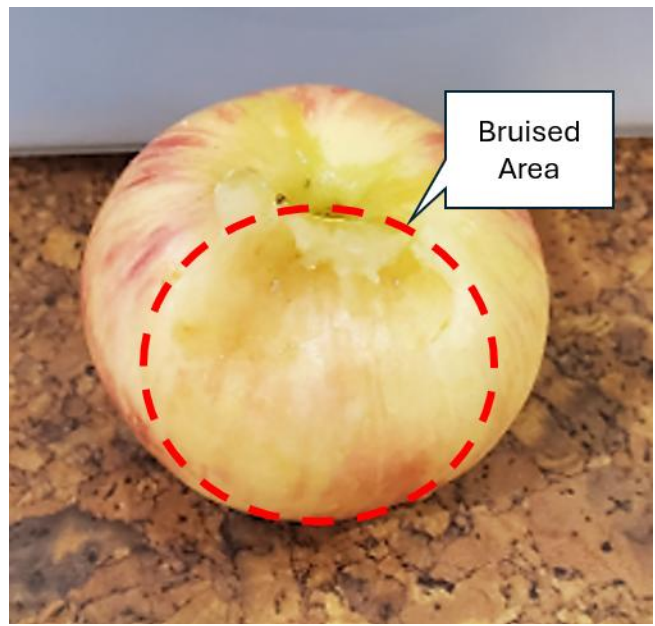

(a)

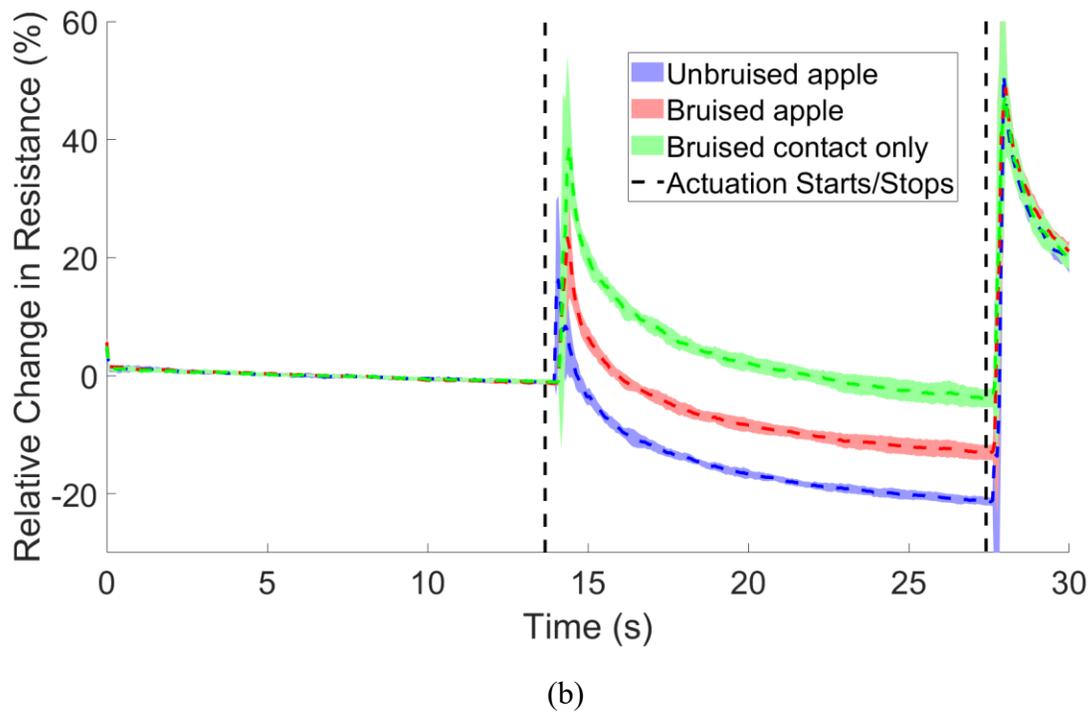

(b)

Figure 17: Experiments on bruised apple. (a) Image of the apple after intentional bruising, (b) mean relative resistance measurements in dashed lines ± 1 standard deviation in shaded region across 10 experiments before and after bruising including 4 pixels only contacting the bruised area.

Table 6: Bruised apple data

| Average resistance change | Before bruising | After bruising | Only bruised |
| --- | --- | --- | --- |
| Exponential estimate | -24.9% ± 0.78% | -15.8% ± 1.3% | -5.89 ± 3.7% |
| Measured resistance at t=20s | -21.2% ± 1.9% | -13.0% ± 2.4% | -3.75 ± 1.6% |

## 5. Conclusion

In this work, a flexible, low-cost, and easy-to-fabricate piezoresistive pressure sensor was integrated with both rigid and soft robotic manipulators for grasping of a variety of produce. The sensor is composed of individual pixels that can detect localized pressure induced on the grasped object. A method for quickly estimating the settled resistance using an exponential decay curve was utilized to reduce the necessary measurement time. This enables not only estimation of the grasping force of the object, but also in identifying key characteristics of the object. Specifically, the size and stiffness can be estimated. The grasping force and stiffness estimation were validated using both silicone sheets and soft exercise balls of varying stiffnesses. Estimation of the size, stiffness, and force allows for accurate grasping of a variety of delicate objects ranging from strawberries to apples. These tasks include determining the sizes of strawberries, monitoring the ripening progress of fruits, and examining produce for damage such as bruising or diseases-related soft spots.

Future work on the sensor will include applications for both harvesting and quality control. For harvesting, the sensor could be utilized to verify that proper grasp has been achieved on the produce to harvest it. For quality control, the sensor could be used to monitor for specific diseases in produce that cause variability in stiffness, check ripeness levels to only harvest in-season fruit, as well as to examine the state of items such as meat for properties associated with high quality. Additionally, the sensor could be integrated at the distribution level to automatically select produce of certain ripeness levels, allowing for consumers to select their preference. The effect of noise, as well as the use of filters, both in hardware and software, on the performance of the sensor in more noisy environments would be essential for use in practical agricultural applications. Additionally, specific parameters of the sensing material such as the linearity across larger force ranges and the minimum detection threshold can be pursued for implementation on different robotic grippers with both smaller and larger grasping forces.

Acknowledgements: the work was supported under the MTRAC Program by the State of Michigan 21$^{st}$ Century Jobs Fund received through the Michigan Strategic Fund and administered by the Michigan Economic Development Corporation.